\documentclass[letterpaper, 10 pt, conference]{ieeeconf}

\IEEEoverridecommandlockouts                              

\usepackage{graphicx}
\usepackage{mathptmx} 
\usepackage{amsmath} 
\usepackage{amssymb}  
\usepackage[utf8]{inputenc}
\usepackage[T1]{fontenc}
\usepackage{float}
\usepackage{algorithm}
\usepackage{algpseudocode}
\makeatletter
\let\MYcaption\@makecaption
\makeatother

\usepackage[font=footnotesize]{subcaption}

\makeatletter
\let\@makecaption\MYcaption
\makeatother
\usepackage{booktabs}
\usepackage[nolist,nohyperlinks]{acronym}
\usepackage{siunitx}
\usepackage{makecell}
\usepackage{multirow}
\usepackage{url}
\usepackage{cite}
\usepackage{tikz}
\usepackage{hyperref}
\usetikzlibrary{arrows.meta,backgrounds,fit}
\newcommand{\FigureTopSpacing}{4.5pt}
\makeatletter
\let\orig@fs@ruled\fs@ruled
\renewcommand\fs@ruled{
  \orig@fs@ruled
  \def\@fs@pre{\kern\FigureTopSpacing\hrule height.8pt depth0pt \kern2pt}
}
\makeatother

\begin{acronym}
  \acro{VBTS}{Vision-Based Tactile Sensor}
  \acro{NN}{Neural Network}
  \acro{ConvLSTM}{Convolutional Long Short Term Memory}
  \acro{CNN}{Convolutional Neural Networks}
  \acro{LSTM}{Long Short Term Memory}
\end{acronym}

\title{\LARGE \bf
Investigating Active Sampling for Hardness Classification with Vision-Based Tactile Sensors}

\author{
Junyi Chen\textsuperscript{1*
}, Frederik Heller\textsuperscript{1*
}, Mario G\'omez Andreu\textsuperscript{1}, Boris Belousov\textsuperscript{2}, Tim Schneider\textsuperscript{1},\\
Lisa P. Y. Lin\textsuperscript{3}, Katja Doerschner\textsuperscript{3}, Knut Drewing\textsuperscript{3}, Jan Peters\textsuperscript{1,2,4,5} and Alap Kshirsagar\textsuperscript{1,6}
\thanks{\textsuperscript{*}Equal contribution}
\thanks{\textsuperscript{1}Intelligent Autonomous Systems Lab, Department of Computer Science, TU Darmstadt, Germany. {\tt\small alap@robot-learning.de}}
\thanks{\textsuperscript{2}German Research Center for AI (DFKI)}
\thanks{\textsuperscript{3}Department of Psychology, University of Giessen, Germany}
\thanks{\textsuperscript{4}Robotics Institute Germany}
\thanks{\textsuperscript{5}Hessian Center for Artificial Intelligence (Hessian.AI)}
\thanks{\textsuperscript{6}Indian Institute of Technology (IIT) Delhi - Abu Dhabi, UAE}
\thanks{This work was supported by the Deutsche Forschungsgemeinschaft (German Research Foundation, DFG) under Germany's Excellence Strategy (EXC 3066/1 “The Adaptive Mind”, Project No. 533717223).}
}

\begin{document}

\maketitle
\thispagestyle{empty}
\pagestyle{empty}

\begin{abstract}
Hardness is one of the most important object properties that humans and robots perceive through touch. This paper investigates active sampling strategies for sample-efficient hardness classification with vision-based tactile sensors (VBTS), using information-theoretic measures of model uncertainty. We evaluated three probabilistic classifier models and two active sampling strategies and a random sampling baseline on a hardness classification task, with our own dataset of VBTS samples collected by a robot and a previously published dataset.
In addition, we conducted a human study on the hardness classification task with the same objects as our dataset.
Our findings indicate that the active sampling approaches, driven by variance and entropy as measures of model uncertainty, perform best with accuracy reaching  $90\%$, compared to human accuracy of $48\%$, demonstrating the effectiveness of using VBTS for hardness classification.
More information can be found at: \url{https://fheller1.github.io/ActiveHardness}
\end{abstract}

\section{Introduction}

Robots are increasingly being utilized in a variety of fields, from manufacturing to healthcare, where they interact with objects in their environment and plan their actions based on sensory feedback. A significant challenge in robotics is accurately perceiving object properties. This work focuses on a crucial property perceived through touch: \emph{hardness}. We investigate active sampling strategies for sample-efficient hardness classification with a \ac{VBTS}. \ac{VBTS} like GelSight Mini~\cite{gelsightmini} or FingerVision~\cite{yamaguchi2019recent} provide a cost-effective and high-resolution state of the art in tactile sensing, leveraging advancements in sensor hardware and computer vision. The ability to rapidly classify an object's hardness can help robots in various tasks like waste sorting, automated harvesting, and handling delicate objects. 
\begin{figure}[ht]
    \centering
    \includegraphics[width=0.98 \linewidth]{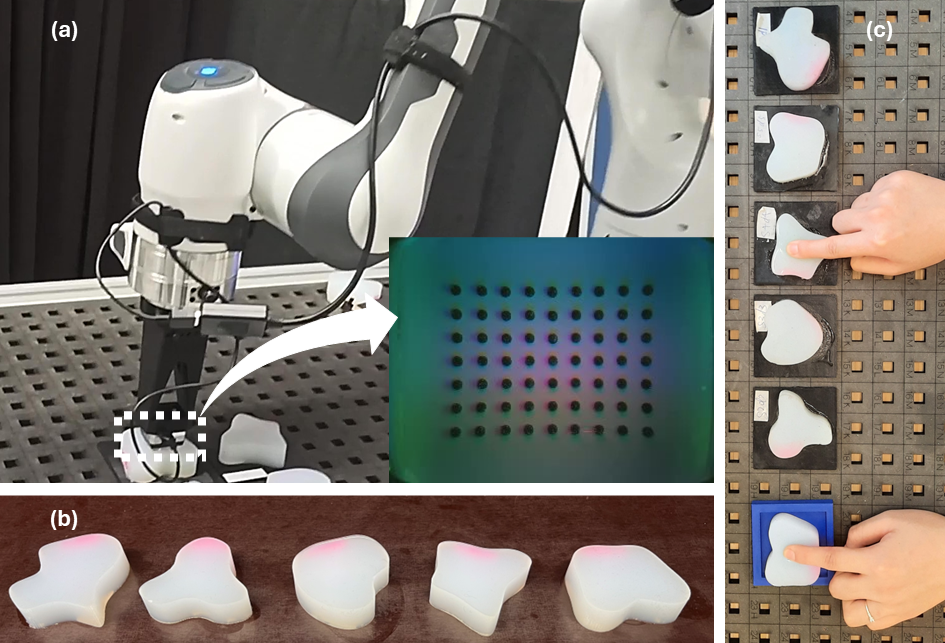}
    \caption{The hardness classification task requires a robot or human to classify which reference object matches a test object in hardness level. (a) The robot uses a GelSight Mini sensor mounted on its end-effector to capture data exploring the hardness of objects. The image captured by the sensor is shown in the inset. (b) Our dataset consists of GelSight Mini videos collected from five silicone objects of different shapes in increasing hardness from left to right. (c) In our human participant study, participants explored the test object (blue plate) and the reference objects (black plates) with their index fingers to classify hardness.}
    \label{fig:setup_dataset}
\end{figure}

Several approaches have been proposed for robotic systems classifying hardness levels or regressing on hardness values. For data collection, these methods rely on deformation measurements by cameras~\cite{Pallottino2011}, force and torque sensors~\cite{burgess2025learning}, or \ac{VBTS} like GelSight~\cite{Yuan2017}. To classify or regress the hardness given sensor measurements, traditional methods used analytical models based on physical rules~\cite{burgess2025learning}, heuristics~\cite{Pallottino2011}, or simple machine learning models like SVM and shallow neural networks~\cite{Zhang2021}. In recent years, deep learning models like \ac{CNN} have become more prevalent. These models can process the high-dimensional data of high-resolution \ac{VBTS} and allow more accurate hardness assessments. Both classification~\cite{Amin2023, Fang2021, Fang2023} and regression~\cite{Nam2024, Chen2022, Yuan2017} models have been proposed in the past. The regression models require true measurements of the hardness values (e.g. using a Shore durometer), which is costly and time-consuming. In contrast, classification models do not require the exact hardness values and avoid measurement bias. Also, classification models can generalize well with fewer training samples. Therefore, in this work, we focus on the task of robotic hardness classification.

The existing hardness classification methods use a fixed number of samples collected from the training classes without regard for sample efficiency. In this work, we explore information-theoretic active sampling strategies for sample-efficient hardness classification. 
The agent is given a \emph{test} object and a number of \emph{reference} objects, and can explore by touching them with a \ac{VBTS}. Without additional prior knowledge about the objects, the task is to classify which reference object is matching the test object in hardness with as few touches of a \ac{VBTS} as possible. 
Previously, Boehm et al.~\cite{alina2024} investigated active sampling strategies for texture recognition with a \ac{VBTS}. In their work, the texture recognition was formulated as an image classification problem. Hardness classification, in contrast, requires processing a sequence of \ac{VBTS} frames, since the hardness becomes apparent in the slight temporal change of the tactile image~\cite{Yuan2017,Fang2021, Fang2023} while pressing on the object. An exemplary input sequence is shown in the first line of Fig.~\ref{fig:model_architectures}.

We provide empirical evidence on the effectiveness of model-uncertainty-based active sampling strategies for hardness classification with a \ac{VBTS}. We first compare three probabilistic classifier architectures on two datasets: our dataset containing videos collected by a robot pressing a GelSight Mini sensor on five silicone objects of different hardness levels (see Fig.~\ref{fig:setup_dataset}) and the dataset created by Yuan et al.~\cite{Yuan2017} containing videos obtained by human testers pressing an older version of the GelSight sensor on silicone objects of different hardness. We investigate two active sampling strategies based on information-theoretic uncertainty measures, and a random sampling baseline. We also conduct a human-participant study to assess the performance of our method vis-à-vis human performance on the hardness classification task on our dataset. 

In the next section, we present our proposed algorithm of active sampling for sample-efficient hardness classification, along with the different classifier model architectures and active sampling strategies.

\section{Method}
For sample-efficient hardness classification using a \ac{VBTS}, our proposed algorithm is outlined in Alg.~\ref{alg:adaptive_sampling}. First, a probabilistic classifier model is trained on a number of initial samples of each reference object. We then make use of a measure of the model uncertainty about each class to decide which reference object is selected to add  additional samples to the training data before retraining the classifier. 
Following, we describe our processing of \ac{VBTS} videos to yield a frame sequence as input for the classifiers, the architectures of our three evaluated probabilistic classifiers, and the two active sampling strategies and the random sampling baseline.

\begin{algorithm}[t]
\caption{Hardness classification with active sampling}\label{alg:adaptive_sampling}
\begin{algorithmic}[1]
      \State \textbf{Input:} Base classes \textit{B}; Initial samples per class $s_0$; Initial training epochs $e_0$; Active sampling iterations $N$; Strategy $S \in \{ \text{``entropy''}, \text{``variance''}, \text{``random''} \}$; New samples introduced to the reservoir per iteration per class $s$; Maximum samples per training iteration $M$; Training epochs per iteration $e$; Predictions per test sample to estimate model uncertainty $n$; Training reservoir \textit{TR}; Training samples \textit{TS}
      \State \textbf{Output:} Trained classifier model

      \State Initialize \textit{TR} with $s_0$ samples from each class in \textit{B}

      \State Train the model for $e_0$ epochs using \textit{TR}

    \For {$i = 1$ to $N$}
         \State Make $n$ predictions on $l$ test samples with dropout

         \If{$S$ $=$ \text{``entropy''}}
            \State Calculate the entropy of predictions (Eq.~\ref{eq:entropy})

            \State \textit{selected class} $\gets$ class with highest entropy 
         \ElsIf{$S$ $=$ \text{``variance''}}
             \State Calculate the variance of predictions (Eq.~\ref{eq:variance})

            \State \textit{selected class} $\gets$ class with highest variance
         \ElsIf{$S$ $=$ \text{``random''}}
            \State \textit{selected class} $\gets$ random class from \textit{B}
         \EndIf

         \State Add $s$ samples from \textit{selected class} to \textit{TR}

         \If{size of \textit{TR} $>$ $M$}
            \State \textit{TS} $\gets$ randomly select $M$ samples from \textit{TR}
         \Else
            \State \textit{TS} $\gets$ \textit{TR}
         \EndIf

         \State Train the model with \textit{TS} for $e$ epochs
      \EndFor
\end{algorithmic}
\end{algorithm}

\subsection{Classifier Input Preparation}
\label{sec:classifier_input}
\begin{figure*}[t]
  \vspace{\FigureTopSpacing}

    \centering
    \setlength{\abovecaptionskip}{3pt}
    \includegraphics[width=\textwidth]{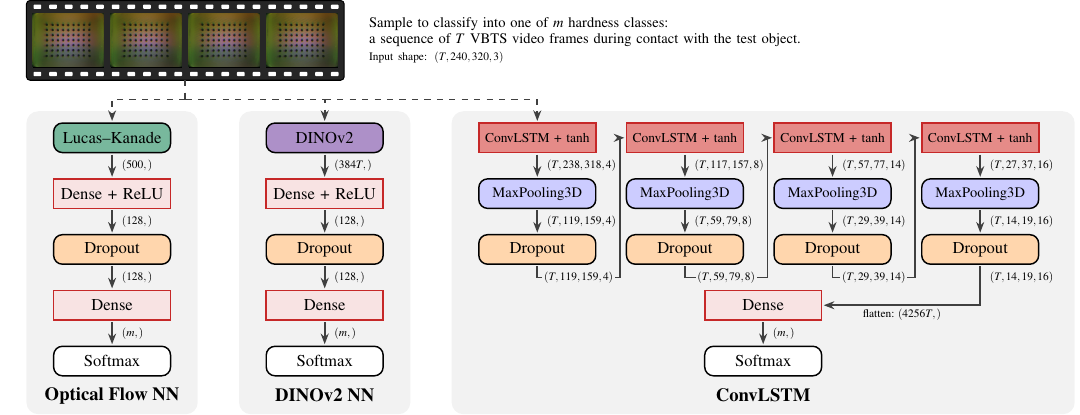}
    \caption{
        Architectures of the three evaluated classification models.
        All models receive video sequences of the VBTS during contact with the test object as input, and predict the hardness class probabilities of the test object over the reference objects. 
        The Optical Flow NN model constructs explicit features using the Lucas-Kanade method~\cite{lucas1981iterative}. In DINOv2 NN, we use a pretrained image feature extractor~\cite{oquab2024dinov2}. The ConvLSTM model implicitly learns features with four sequential blocks of Convolutional LSTM~\cite{shi2015convolutional}, Pooling, and Dropout layers.
    }
    \label{fig:model_architectures}
\end{figure*}

When a \ac{VBTS} like the \emph{GelSight Mini}~\cite{gelsightmini} is pressed against a deformable object, the changes in the color intensity of the image and the displacement of embedded markers (see Fig.~\ref{fig:model_architectures}) on the gelpad encode information about the object's hardness~\cite{Yuan2016}. To obtain a fixed-length sequence of $T$ frames for different presses, we first constrain the video to the loading period of the press, i.e. from the point of initial contact to the point of maximum deformation. This helps to make the methods robust to the pressing motion's speed and the maximum contact force, which in general differs between different presses. We subtract the initial frame in the video sequence from every frame to account for any preexisting elastomer deformations. Then, we determine the initial contact point by identifying the first frame where the mean intensity of the image is beyond a threshold, and the point of maximum deformation as the frame where the intensity difference reaches its peak. We then select $T-2$ additional frames equally spaced between the initial contact and maximum deformation frame to yield our input sequence.

\subsection{Model Architectures}
\label{sec:classifier_models}
For the task of hardness classification given an input sequence prepared from a \ac{VBTS} video, we implement three different probabilistic classifier models. An input sequence and the architectures of each classifier are shown in Fig.~\ref{fig:model_architectures}.
Common to all proposed models is a classification head consisting of a fully connected (dense) \ac{NN}, followed by the Softmax operator to obtain a probability distribution over the $m$ reference object hardness classes. Dropout layers are used in each architecture to estimate the classifier's uncertainty: By randomly deactivating a subset of neurons during inference, dropout effectively simulates an ensemble of models, providing a measure of the network's uncertainty through the distribution of predictions~\cite{gal2016dropout}. The models differ in how features are constructed before the classification head.

\subsubsection*{Model 1 -- Optical Flow Features and Neural Network Classifier (OFNN)}
For the OFNN model, we explicitly construct key motion features from the image sequence. Since object hardness influences the way the gel deforms when it is pressed against the object, optical flow features help capture these deformations by tracking motion patterns in the video, specifically of the black markers. The model uses $500$ Lucas-Kanade optical flow features~\cite{lucas1981iterative} extracted from the input sequence. Then, a two-layer \ac{NN} with 128 neurons in the hidden layer and a dropout layer is used to classify the hardness. 

\subsubsection*{Model 2 -- DINOv2 Features and Neural Network Classifier (DINOv2NN)}
The DINOv2NN model employs the pretrained  vision foundation model \emph{DINOv2}~\cite{oquab2024dinov2} to extract versatile visual features from images. We concatenate the features extracted from each frame in the sequence. The concatenated feature vector is fed to a simple two-layer \ac{NN} classifier, as also used in the OFNN model. We use a pretrained \emph{DINOv2-S/14} backbone and only update the weights of the \ac{NN} classifier during training.

\subsubsection*{Model 3 -- Convolutional Long-Short Term Memory Network Classifier (ConvLSTM)}
The \ac{ConvLSTM} architecture, proposed by Shi et al.~\cite{shi2015convolutional}, combines the strengths of a \ac{CNN} and a \ac{LSTM} neural network. The \ac{CNN} layers extract spatial features from each frame, while the \ac{LSTM} component learns and retains temporal relationships during the sequence. Due to its seamless compatibility with sequential input, the ConvLSTM architecture was successfully used in previous work~\cite{Yuan2017, Chen2022} for hardness estimation with \ac{VBTS}. Our implementation of the model consists of four sequential blocks of a ConvLSTM with 4, 8, 14, and 16 filters followed by Tanh activation, pooling, and a dropout layer, followed by a single dense layer in the classification head. This model implicitly constructs useful features from the input sequence in the ConvLSTM blocks to classify object hardness.

\subsection{Active Sampling Strategies}
\label{sec:active_sampling_strategies}
As shown in Alg.~\ref{alg:adaptive_sampling}, the agent first trains a classifier model on a number of initial samples per reference object. Following this initial training, the agent decides on the next reference object based on the model uncertainty, of which more samples are collected and added to the training data. To decide on the next sampled reference object, the agent queries the model repeatedly with the same test samples and estimates the uncertainty from the distribution of the predictions generated due to the dropout layers. Similar to Boehm et al.~\cite{alina2024}, we compare two information-theoretic uncertainty metrics---\textit{entropy} and \textit{variance}---obtained from the distribution of classifier predictions given the test samples.

Where $p_{i, j, k}$ is the probability of the model classifying sample $k \in \{1, ..., l\} $ in the test set as class $i$ in the $j$-th query ($i \in \{1, ..., m\};j \in \{1, ..., n\}$),
the entropy expectation of a class $i$ is given by
\begin{equation}
	\begin{split}
		H_i &= -\displaystyle \mathop{\mathbb{E}}_{j, k}  (p_{i, j, k} \log(p_{i, j, k})) = -\frac{1}{nl} \mathop{\sum}_{k=1}^{l} \mathop{\sum}_{j=1}^{n}  (p_{i, j, k} \log(p_{i, j, k})) \text{.}
	\end{split}
 \label{eq:entropy}
\end{equation}
Similarly, variance is defined as 
\begin{equation}
	\begin{split}
		Var_i &=\displaystyle \mathop{\mathbb{E}}_{j, k}  (p_{i, j, k} - \mu_i)^2 = \frac{1}{nl} \mathop{\sum}_{k=1}^{l} \mathop{\sum}_{j=1}^{n}  (p_{i, j, k} - \mu_i)^2 \text{,}
	\end{split}
 \label{eq:variance}
\end{equation}
where $\mu_i$ is the mean of the $n$ predictions of class $i$, given by
\begin{equation}
    \mu_i = \frac{1}{nl} \mathop{\sum}_{k=1}^{l} \mathop{\sum}_{j=1}^{n} p_{i, j, k} \text{.}
\end{equation}
The agent selects the reference object with the highest uncertainty as measured by entropy or variance and collects a number of additional samples from that object to refine the classifier model. As an additional baseline, we implement a \emph{random} sampling strategy, in which the agent selects the next reference class randomly sampled from a uniform distribution. 

\subsection{Retraining with Reservoir Sampling}
To manage the size of the training data, we employ reservoir sampling to select a subset of samples from the growing dataset. The agent adds newly collected reference samples to the training reservoir and retrains the classifier model using the updated training reservoir. If the size of the training reservoir exceeds a predefined threshold $M$, the agent randomly selects a subset of size $M$ from the reservoir for training. Otherwise, the agent uses all available samples for training.

\section{Evaluation}
We evaluate the effectiveness of our active sampling algorithm for sample-efficient hardness classification, and compare its performance with human performance. To do so, we first evaluate the three classifier models from Sec.~\ref{sec:classifier_models} on our dataset and a dataset previously published by Yuan et al.~\cite{Yuan2017} without resampling. Then, we test the active sampling strategies with the best model for each dataset. Finally, we conduct a human-participant study to estimate the hardness classification accuracy of humans on the objects in our dataset.

\subsection{Datasets}
\label{sec:datasets}
We tested our hardness classification method on two datasets: a dataset that we created containing five hardness classes and a dataset created by Yuan et al.~\cite{Yuan2017} containing nine hardness classes.

\subsubsection{Our Dataset}
\label{sec:our_dataset}
We created a set of objects, shown in Fig.~\ref{fig:setup_dataset}, of five different hardness levels by mixing a two-component silicone rubber solution (Alpa Sil EH A and B) with different amounts of silicone oil. The resulting hardness levels were \SI{1.13}{mm \per N}; \SI{1.02}{mm \per N}; \SI{0.79}{mm \per N}; \SI{0.68}{mm \per N}; \SI{0.44}{mm \per N}, listed from softest to hardest. We used 3D-printed molds to cast the objects into shapes, ensuring flat surfaces free of air pockets and any noticeable texture differences.

To collect data of the robot exploring the hardness of these objects, we mounted a \emph{GelSight Mini}~\cite{gelsightmini} tactile sensor on the end effector of a Franka Emika \emph{Panda} $7$-DOF robot arm. The robot pushed the tactile sensor on the top surface of each object, as shown in Fig.~\ref{fig:setup_dataset}, with a given velocity and retreated when the contact force as measured by the Panda joint torque sensors reached a given threshold, while recording videos from the GelSight's internal camera. The contact force was used solely for variance in data collection, not for hardness classification. The robot always pushed the sensor in the center of the object so that the object shape was not captured, and the hardness classification would not be influenced by shape or texture.

Each captured push and video represents one data point, for a total of $455$ recordings in our dataset. To increase the variance of the data on each object, the maximum contact force and the downward velocity of the end-effector were randomly sampled from uniform distributions,  with contact force between $[\SI{1}{\newton}, \SI{5}{\newton}]$ and velocity in $[\SI{20}{mm \per s}, \SI{50}{mm \per s}]$. Each captured video is labeled the hardness class of this object in $\{1,\ldots, 5\}$. The dataset also contains the contact force between the \emph{GelSight} sensor and the object, measured by the robot's joint torque sensors, as well as a timestamp for each frame in the videos. Our dataset is released on the website: \url{https://fheller1.github.io/ActiveHardness}.
We apply the procedure described in Sec.~\ref{sec:classifier_input} to obtain a classifier input sequence of length $T=4$ from each video.

\subsubsection{Yuan Dataset}
\label{sec:yuan_dataset}
The dataset published by Yuan et al.~\cite {Yuan2017} comprises approximately $7000$ videos, each representing an independent video of the \textit{GelSight} sensor pressed by a human tester onto one of silicone samples of different hardness. Their dataset is annotated with the true Shore $00$ scale hardness of each object. For our experiments, we select the ``flat'' object category, and Shore $00$ hardnesses of $8$, $17$, $27$, $46$, $62$, $73$, $81$, $87$, and $95$. Due to varying video length, we process the videos into input sequences of length $T=3$ to ensure at least 10 available samples per object.

\subsection{Experiment Settings}
\begin{figure}[t]
  \vspace{\FigureTopSpacing}

    \centering
    \setlength{\abovecaptionskip}{3pt}
    \includegraphics[width=\linewidth]{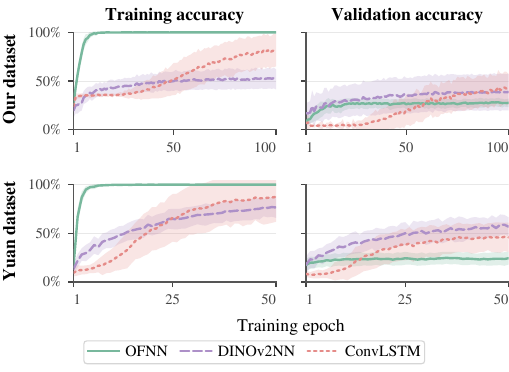}
    \caption{Training and validation accuracy of the proposed OFNN, DINOv2NN and ConvLSTM classifier models trained without resampling on our dataset and the Yuan dataset. Each curve shows the mean $\pm$ one standard deviation over 50 runs per model and dataset.}
    \label{fig:model_baselines}
\end{figure}

The number of available hardness classes is $m=5$ in our dataset, and $m=9$ in our selected subset of the Yuan dataset.
We set the number of initial samples collected from each reference object and the number of new samples collected in each active sampling iteration to five. Initial experiments showed that the samples in our dataset were harder to classify than the samples in the Yuan dataset (see Fig.~\ref{fig:model_baselines}). Therefore, for our dataset, we set the number of initial training epochs to $100$, with $50$ additional epochs per active sampling iteration. For the Yuan dataset, we set the initial training epochs to $50$, with $25$ additional epochs per active sampling iteration. Due to GPU memory constraints, we restricted the number of training samples selected from the reservoir per iteration to $M=80$ for the OFNN and DINOv2NN models, and to $M=40$ for the ConvLSTM model, which required significantly more GPU memory due to its high number of trainable weights.

\subsection{Evaluation Metrics}
After the initial training, we evaluate the model's accuracy on test samples with the dropout layer disabled during inference. We also assess the model's accuracy after each active sampling iteration to monitor its performance throughout the active sampling process. Our models operate as classifiers rather than as regressors, making them agnostic to the specific hardness metric of the samples. Nevertheless, as our hardness classes are ordered, a misclassification of a class closer to the true hardness class may be preferable.
Therefore, alongside accuracy, we compute the Mean Absolute Error (MAE) between the predicted and actual hardness class values in \SI{}{mm \per N} to assess the model's pseudo-regression and ordinal classification capabilities of quantifying the hardness level. For the Yuan dataset, the MAE can be directly computed on the known Shore hardnesses.  

\subsection{Model Baselines}
\label{sec:model_baselines}
We first compare the performance of the three classifier models—\textit{OFNN}, \textit{DINOv2NN}, and \textit{ConvLSTM}—trained with the initial samples without any active sampling on both datasets. Each model is trained on five samples per class and evaluated on five validation samples per class. Fig.~\ref{fig:model_baselines} shows the training and validation accuracies of the models for each dataset, averaged over $50$ runs per model and dataset with different random data sampling. We find that \textit{ConvLSTM} outperforms the other two models on our dataset, achieving a validation accuracy of $41.76\%$, \textit{DINOv2NN} follows close behind with an accuracy of $38.80\%$, and \textit{OFNN} performs the worst with a validation accuracy of only $27.60\%$. On the Yuan dataset, \textit{DINOv2NN} outperforms the other two models, achieving a validation accuracy of $56.85\%$, \textit{ConvLSTM} is next with an accuracy of $45.90\%$, and \textit{OFNN} performs the worst with a validation accuracy of only $24.40\%$.

\subsection{Ablation Study on Dropout Rate}
\begin{figure}[t]
  \vspace{\FigureTopSpacing}

    \centering
    \setlength{\abovecaptionskip}{3pt}
    \includegraphics[width=\linewidth]{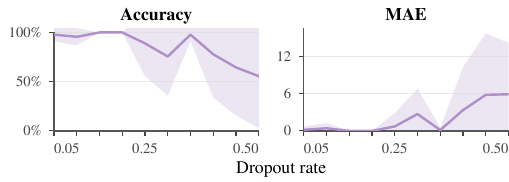}
    \caption{Effect of dropout rate on accuracy and MAE on the Yuan dataset with the DINOv2NN model. Each curve shows the mean $\pm$ one standard deviation over $9$ runs with different test objects.}
    \label{fig:drop_acc_MAE}
\end{figure}

As described in Sec~\ref{sec:classifier_models}, we use dropout layers to estimate the model uncertainty. If the dropout rate is too high, the uncertainty evaluation procedure may suffer from randomness and subsequent inability to estimate the model uncertainty. To test the effect of different dropout rates, we evaluated the performance of the DINOv2NN model on the Yuan dataset with ten different dropout rates spaced equally from $0.05$ to $0.5$.
For each tested dropout value, we trained $9$ DINOv2NN models, one per test object in the Yuan dataset, and evaluate their performance after $5$ active sampling updates with the variance strategy. Fig.~\ref{fig:drop_acc_MAE} plots the mean and standard deviation of accuracy and MAE for each dropout rate.
We find that a dropout rate higher than $0.2$ causes a significant degradation in the performance and stability of the model. For the dropout rate $0.35$, the accuracy is almost as high as for the lower dropout rates, which may be explained by favorable randomly sampled data. In our following evaluation of the active sampling strategies, we fix the dropout rate to $0.2$.

\subsection{Active Sampling Performance}
\label{sec:active_sampling_performance}
We test both information-theoretic active sampling strategies (\textit{entropy} and \textit{variance}) and compare them to the \textit{random} sampling baseline on our dataset and the Yuan dataset. For each dataset, we select the best-performing classifier model as determined in Sec.\ref{sec:model_baselines} for the evaluation, i.e., \textit{ConvLSTM} for our dataset and \textit{DINOv2NN} for the Yuan dataset.

\begin{figure*}[t]
  \vspace{\FigureTopSpacing}

    \centering
    \setlength{\abovecaptionskip}{3pt}
    \includegraphics[width=\textwidth]{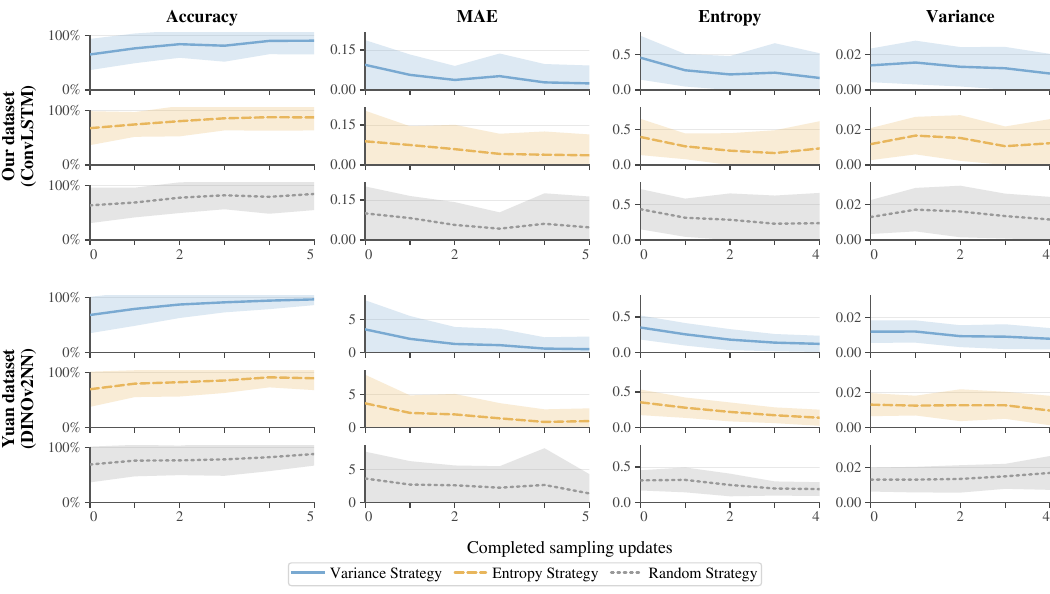}
    \caption{Comparison between active sampling strategies and the random sampling baseline for the ConvLSTM model on our dataset and the DINOv2NN model on the Yuan dataset.
    Shown are mean and standard deviation of the metrics accuracy, MAE, entropy and variance over the number of completed active sampling updates. Since entropy and variance metrics were measured only before an active sampling update in our experiments, they are not measured after the final active sampling update.}
    \label{fig:active_sampling_results}
\end{figure*}

\subsubsection{Our Dataset}
The \textit{ConvLSTM} model is initially trained on $5$ samples per class, and then we run the active sampling algorithm for $5$ iterations. Each iteration adds $5$ samples from the selected class to the training reservoir. The results are averaged over $5$ comparison classes for the test object, where each class is replicated with $10$ runs. Therefore, there are $50$ runs for each strategy, for a total of $150$ runs. 

We observed that of $150$ original runs on the \textit{ConvLSTM} model, $28$ runs diverged with a training loss greater than $1.0$ after training on the initial samples, and low classification accuracy due to mostly constant predictions without regard for the input. This divergence is likely caused by the relatively high complexity of the \textit{ConvLSTM} model combined with the relatively small amount of training data. Thus, additional runs were performed to obtain $150$ converged models as planned.

The top three rows of Fig.~\ref{fig:active_sampling_results} show the mean accuracy and MAE for each of the three strategies on our dataset over the number of completed active sampling updates. The metrics after $0$ updates refer to the model after training on the initial samples. Since the entropy and variance metrics are computed before each active sampling update, there is no measure for these metrics after the final update.
We find that both active sampling strategies outperform the baseline \emph{random} strategy in accuracy and MAE, with the \emph{variance} strategy slightly improving on the \emph{entropy} strategy in most active sampling iterations. The mean variance and entropy metrics decrease roughly monotonically with the variance strategy. The entropy strategy and random baseline seem to slightly decrease the mean entropy with more sampling updates, while the mean variance does not clearly follow this trend. However, the standard deviation bands around all metrics are relatively large. 

\subsubsection{Yuan Dataset}
Similarly, the \textit{DINOv2NN} model is initially trained on $5$ samples per class, then the \textit{active sampling} algorithm is applied for $5$ iterations. We add $5$ samples from the selected class in each iteration. The results are averaged over $9$ comparison classes, where each class is replicated with $5$ runs. Therefore, there are $45$ runs for each strategy for a total of $135$ runs. We did not observe any convergence issues with the \textit{DINOv2NN} model on the Yuan dataset. 

The bottom three rows of Fig.~\ref{fig:active_sampling_results} show the mean accuracy, MAE, entropy, and variance for the active sampling strategies on the Yuan dataset. Again, we  find that both active sampling strategies outperform the random baseline.
Particularly, the variance strategy achieves very high mean accuracy with low standard deviation. The mean MAE is comparably low with the variance and entropy strategy, where the random baseline shows higher mean MAE. The entropy decreases relatively similarly with updates in all three strategies. The variance slightly decreases in the variance and entropy strategies, while it slightly increases with updates in the random baseline. The standard deviation bands for the metrics are also large.

To investigate whether the variance and entropy strategies beat the random baseline with statistical significance, we perform one-sided permutation tests on accuracy and MAE for each strategy. The groups were shuffled with $99,999$ permutations within their reference object classes, and $p$-value results adjusted by Holm correction for the four tests per dataset. The variance strategy significantly beats the random baseline in accuracy on the Yuan dataset, with an adjusted $p$-value of $0.027$. For the other strategies, metrics, and dataset, no statistical significance can be claimed.

\subsection{Human Study and Comparison}
\label{sec:human_study}
We conducted a human-participant study to investigate and compare to the human performance in hardness classification on the objects in our dataset. We recruited $n=10$ participants from the student population at TU Darmstadt for the study. All participants reported no issues with their tactile perception ability. Each experiment trial consisted of a practice round and five test rounds. In each round, the participants were given our five reference objects of different hardness, and a test object that matched one of the reference objects in hardness.
The participants were blindfolded and allowed to press the objects only with their index fingers to keep the setup similar to the robotic setup. The participants pressed the test object with their left index finger while exploring the reference objects with their right index finger. 

Each round had a time limit of 30 seconds, which was determined after pilot trials showing that the duration was enough for participants to answer which reference object matches the test object's hardness. Each participant could also end a round earlier by verbally confirming their ``classification'' choice. In each of the five test rounds, we used a different test object from our dataset (see Sec.~\ref{sec:our_dataset}) and shuffled the sequence of the five reference objects.

Table~\ref{table:experiment_acc} compares the human performance in our human study with the performance of our models trained in Sec.~\ref{sec:active_sampling_performance}. For the human performance, we compute mean and standard deviation over $10$ participants. For the ConvLSTM model on our dataset, we give the mean and standard deviation over all $150$ runs before any resampling (No resampling). For the active sampling strategies and random baseline after $5$ active sampling updates, metrics are computed over $50$ runs. Similarly, we compare the performance of the DINOv2NN model on the Yuan dataset with all $135$ runs for no resampling, and $45$ runs each for active sampling strategies. For the Yuan dataset, there is no human comparison since the physical objects were not available to us.

As Table~\ref{table:experiment_acc} shows, the mean classification accuracy in the human setting on our dataset over all participants and rounds was $48\%$ with a standard deviation of $23.5\%$. The MAE on the classified hardness values was \SI{0.103}{mm \per N} with a standard deviation of $0.049$. All compared ConvLSTM model ablations outperform the human result. The models without resampling improve on the human performance, with a mean of $65\%$ accuracy and \SI{0.094}{mm \per N} MAE, although slightly higher standard deviation. The active sampling strategies and random baseline are close together in performance, with the variance strategy scoring the highest mean accuracy of $90\%$ and lowest MAE of \SI{0.025}{mm \per N}, and the entropy strategy and random baseline achieving mean accuracy of around $87\%$ and $84\%$, respectively. On the Yuan dataset with the DINOv2NN model, the variance strategy similarly takes the lead with a mean accuracy of almost $97\%$ and a MAE of $0.507$ on the Shore $00$ scale. The entropy strategy and random baseline achieve mean accuracies of around $90\%$ and $88\%$, clearly beating the performance without resampling, which is $69\%$ accurate in its mean.

\begin{table}[t]
  \vspace{\FigureTopSpacing}

\centering
\caption{
Comparison of hardness classification performance between human participants and our methods, measured by mean accuracy (Acc.) and MAE, each metric $\pm$ one STD.
}
\label{table:experiment_acc}
\begingroup
\footnotesize
\setlength{\tabcolsep}{2pt}
\renewcommand{\arraystretch}{1.2}
\newcommand{\tablevalue}[3]{\makebox[.8em][r]{$#1$}\makebox[1em][r]{$#2$}$.$\makebox[1.5em][l]{$#3$}}
\newcommand{\besttablevalue}[3]{\makebox[.8em][r]{$#1$}\makebox[1em][r]{\textbf{#2}}\textbf{.}\makebox[1.5em][l]{\textbf{#3}}}
\begin{tabular*}{\linewidth}{@{}c@{\extracolsep{\fill}}lccccc@{}}
 \toprule
 & & Humans & \makecell{No\\resampling} & Variance & Entropy & Random \\
 \midrule
 \multirow{4}{*}{\rotatebox{90}{\makecell{Ours\\ConvLSTM}}}
 & \multirow{2}{*}{Acc.} & \tablevalue{}{48}{0} & \tablevalue{}{65}{1} & \besttablevalue{}{90}{0} & \tablevalue{}{86}{8} & \tablevalue{}{84}{0} \\
 & & \tablevalue{\pm}{23}{5} & \tablevalue{\pm}{30}{5} & \tablevalue{\pm}{25}{0} & \tablevalue{\pm}{23}{8} & \tablevalue{\pm}{29}{4} \\
 & \multirow{2}{*}{MAE} & \tablevalue{}{0}{103} & \tablevalue{}{0}{094} & \besttablevalue{}{0}{025} & \tablevalue{}{0}{036} & \tablevalue{}{0}{047} \\
 & & \tablevalue{\pm}{0}{049} & \tablevalue{\pm}{0}{103} & \tablevalue{\pm}{0}{067} & \tablevalue{\pm}{0}{078} & \tablevalue{\pm}{0}{116} \\
 \midrule
 \multirow{4}{*}{\rotatebox{90}{\makecell{Yuan\\DINOv2NN}}}
 & \multirow{2}{*}{Acc.} & \multirow{2}{*}{---} & \tablevalue{}{69}{2} & \besttablevalue{}{96}{9} & \tablevalue{}{89}{8} & \tablevalue{}{88}{4} \\
 & & & \tablevalue{\pm}{32}{4} & \tablevalue{\pm}{10}{4} & \tablevalue{\pm}{21}{6} & \tablevalue{\pm}{21}{1} \\
 & \multirow{2}{*}{MAE} & \multirow{2}{*}{---} & \tablevalue{}{3}{607} & \besttablevalue{}{0}{507} & \tablevalue{}{0}{973} & \tablevalue{}{1}{378} \\
 & & & \tablevalue{\pm}{4}{253} & \tablevalue{\pm}{1}{915} & \tablevalue{\pm}{1}{935} & \tablevalue{\pm}{2}{938} \\
 \bottomrule
\end{tabular*}
\endgroup
\end{table}

\begin{figure}[t]
  \vspace{\FigureTopSpacing}

    \centering
    \setlength{\abovecaptionskip}{3pt}
    \includegraphics[width=\linewidth]{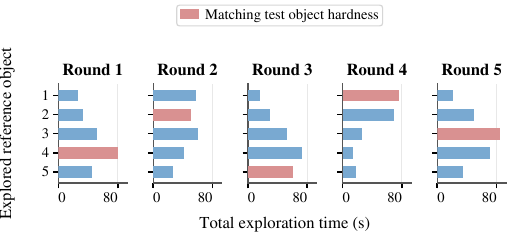}
    \caption{Exploration time per reference object, summed across all human participants, for each round. Red bars mark the reference object matching the test objects hardness in this round.}
    \label{fig:total_time_per_obj}
\end{figure}

Finally, we annotated videos of the human experiments with the positions of fingers and objects, using a custom semi-automatic pipeline, consisting of MediaPipe for hand tracking~\cite{lugaresi2019mediapipe} and Segment Anything Model-2~\cite{ravi2025sam} for object detection.
In this way, we identified frames in which the right index finger touched a reference object, and then counted the time that participants spent exploring each object during an experiment round. Fig.~\ref{fig:total_time_per_obj} shows, for each round, the time spent on each reference object summed over all $10$ participants. It can be seen that participants relatively quickly exclude reference objects whose hardness is far from the test object from the comparison, and instead spend more time exploring the reference objects that are similar to the test objects hardness. Misclassifications also mostly occurred with reference objects that were close to the correct reference object, illustrated by the rather low accuracy but also low MAE for humans in Table~\ref{table:experiment_acc}.

\section{Discussion and Conclusion}
In this work, we proposed and investigated a model uncertainty-based approach to active sampling for sample-efficient hardness classification with a \ac{VBTS}. We evaluated different classifier architectures, dropout rates, and active sampling strategies to find what matters for high classification accuracy and low MAE on both a preexisting dataset, and our own dataset, which we release along with our findings. We also performed a comparative human participant study with similar settings.

First, we compared three different architectures for a probabilistic classifier for hardness classification. We evaluated their performance after initial training without any resampling on the two datasets.
On our dataset, the best mean accuracy of $41.76\%$ was achieved by the ConvLSTM model, and on the Yuan dataset, the DINOv2NN model achieved the highest mean accuracy with $56.85\%$ (Fig.~\ref{fig:model_baselines}).
Comparing to the approach by  Boehm et al. in object texture recognition using \ac{VBTS}~\cite{alina2024}, their classifier achieved $80\%$ accuracy even with just one sample per class, compared to five samples per class in our experiments. 
Additionally, contrasting with the findings of Boehm et al., where rotation-based data augmentation greatly improved the model performance, we found rotation-based data augmentation techniques ineffective for our task. These results suggest that hardness classification is more difficult than texture recognition for a \ac{VBTS}. Therefore, an investigation of the active sampling strategies from~\cite{alina2024} was warranted for the task of hardness classification.

Second, we investigated two active sampling strategies aimed at maximizing information gain by selecting samples with higher model uncertainty, and compared to a baseline where the sampled class was selected randomly. Over five active sampling updates, the variance strategy achieved the highest mean accuracy and lowest MAE on both datasets (Fig.~\ref{fig:active_sampling_results}). A clear, statistically significant advantage of the active sampling strategies over the random sampling baseline could however be shown only for the Yuan dataset. Still, as the models learn from more samples, every strategy shows a decrease in the entropy. In contrast, reducing the variance seems to be harder, and only the \textit{variance} strategy reduces it effectively.

Third, we conducted a human-participant experiment to compare the accuracy of our \ac{VBTS}-based hardness classification approaches to the accuracy of humans on the same set of objects. We found that a \ac{ConvLSTM} classifier trained on only initial samples already achieved around $17$ percentage points higher mean accuracy than human participants. With $5$ active sampling updates, both active sampling strategies and the random baseline further improved the accuracy, reaching up to $90\%$ accuracy on our dataset, which beats the human performance by over $40$ percentage points (Table~\ref{table:experiment_acc}). The low accuracy in human hardness classification could be attributed to confusing similar hardness values of the reference objects (Fig.~\ref{fig:total_time_per_obj}). 
Our findings demonstrate the effectiveness of \ac{VBTS}-based hardness classification compared to humans.

\emph{Limitations and Future Work:} 
While the comparison with human participants on the hardness classification task on our dataset generally illustrates the capabilities of our proposed methods, we recognize that a fair comparison is hard. While parameters like the time limit for humans or the number of samples used for training the models were chosen to achieve a roughly comparable experiment setting, they are still somewhat arbitrary and influence the performance. Also, while the ConvLSTM model finally achieved much better classification accuracy than humans on our dataset, model runs that diverged had to be excluded.
The large standard deviation bands likely reflect differences in classification difficulty across hardness classes in both datasets, which is consistent with the high variance observed in the human participant study.

Finally, our experiments only considered the hardness classification of flat objects, whereas several real-world hardness classification tasks involve complex object shapes. Also, the robot and the human participants were restricted to single-finger explorations of the objects. 
Future work could focus on sample-efficient hardness classification of complex objects with multi-modal data such as proprioceptive, visual, and tactile data. Another direction of future work can investigate human-inspired active sampling as well as active exploration strategies with single and multi-fingered grippers.

\bibliographystyle{IEEEtran}
\bibliography{reference}

\end{document}